\documentclass[letterpaper, 10 pt, conference]{ieeeconf}  

\IEEEoverridecommandlockouts                              

\usepackage{graphics} 
\usepackage{epsfig} 
\usepackage{times} 
\usepackage{amsmath} 
\usepackage{amssymb}  
\usepackage{bm}
\usepackage[nolist]{acronym}
\usepackage{cite}
\usepackage{units}
\usepackage{microtype}
\usepackage{booktabs}
\usepackage[hyperfootnotes=false]{hyperref}

\newcommand{\etal}{\textit{et al}.}

\title{\LARGE \bf
Local and Global Information in Obstacle Detection on Railway Tracks
}

\author{Matthias Brucker$^{2,\ast}$, Andrei Cramariuc$^{1,\ast}$, Cornelius von Einem$^{1,\ast}$, Roland Siegwart$^{1}$, and Cesar Cadena$^{1}$
\thanks{$^\ast$Authors contributed equally to this work, ordered alphabetically.}%
\thanks{$^1$Authors are members of the Autonomous Systems Lab, ETH Zurich, Switzerland; {\tt\small \{firstname.lastname\}@mavt.ethz.ch}}%
\thanks{$^2$Author is with Magazino, Germany but the work was done while the author was a member of $^1$;}%
\thanks{This work was supported by the ETH Mobility Initiative under the project \textit{LROD-ADAS}. The code is available at \url{https://github.com/ethz-asl/railway-anomaly-detection}}%
}
\begin{document}

\maketitle
\thispagestyle{empty}
\pagestyle{empty}

\begin{abstract}
Reliable obstacle detection on railways could help prevent collisions that result in injuries and potentially damage or derail the train. Unfortunately, generic object detectors do not have enough classes to account for all possible scenarios, and datasets featuring objects on railways are challenging to obtain. We propose utilizing a shallow network to learn railway segmentation from normal railway images. The limited receptive field of the network prevents overconfident predictions and allows the network to focus on the locally very distinct and repetitive patterns of the railway environment. Additionally, we explore the controlled inclusion of global information by learning to hallucinate obstacle-free images. We evaluate our method on a custom dataset featuring railway images with artificially augmented obstacles. Our proposed method outperforms other learning-based baseline methods.
\end{abstract}

\section{INTRODUCTION}

With rising global demand for transportation by rail, both due to increasing global trade and changing consumer behavior, railway networks are reaching their operational capacities. 
To ensure the safe operation of trains on increasingly busier tracks, upgrades to the railway control systems are required in the form of reliable communication, continuous and accurate localization~\cite{ETCSEngineers2011}, and environmental awareness in the form of obstacle detection systems.
Life-threatening risks to humans and extensive disruptions to the railway network operation can be prevented through the reliable detection of humans, animals or any unknown objects on the train tracks.

\begin{figure}
    \centering
    \includegraphics[width=\linewidth]{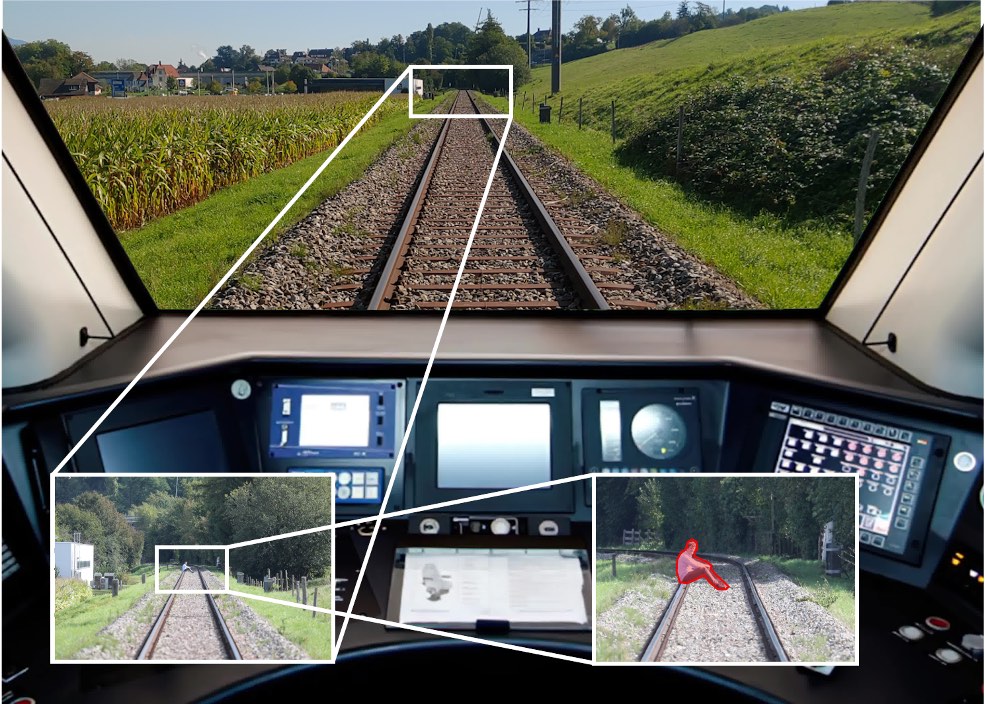}
    \caption{Concept of a zoomable view enhancement with obstacle detection.}
    \label{fig:new_teaser}
\end{figure}

High vehicle speeds, low braking forces, and the great weight of trains result in braking distances exceeding several hundreds of meters, out of the range of common obstacle detection sensor systems, such as LiDAR or stereo vision. 
We thus propose a novel active long-range obstacle detection system, consisting of a zoomable high-focal length camera on an actuated platform~\cite{voneinem2022}, to detect potential obstacles, even at great distance~\cite{ukaiObstacleDetectionSequence2004} as shown in Figure~\ref{fig:new_teaser}.
A critical aspect of such a system is the accuracy of the visual detection of known and unknown entities on the railway.
So far, most research has focused on detecting pre-defined categories, such as humans or other trains, by training object detection networks on custom railway obstacle datasets~\cite{Yu2018,liRealWorldRailwayTraffic2018,Xu2019,Chernov2020,Ristic-Durrant2021a}. 
These systems are, however, by design limited to this pre-defined set of categories and fail to generalize to unknown obstacle types. 
Moreover, the models must be trained on datasets containing artificially inserted obstacles or datasets out of the railway domain, as real images containing obstacles on rails are challenging to obtain.
This leads to strong biases in the detectors and raises doubts about their capabilities in real-world applications.

\begin{figure}
    \centering
    \includegraphics[width=0.49\linewidth]{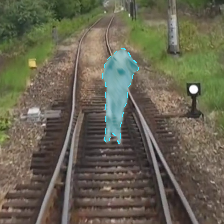}
    \includegraphics[width=0.49\linewidth]{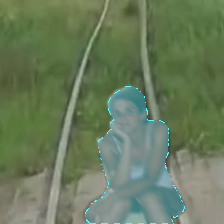}
    \caption{Railway obstacle detection on the \textit{FishyRails} dataset. Annotated in blue are the synthetically placed obstacles and our detection of them.}
    \label{fig:teaser}
\end{figure}

In this work, we focus on the task of data-driven anomaly detection, where instead of learning to detect a predefined set of obstacle categories, we learn to recognize the railway environment and by exclusion everything that does not belong, \textit{i.e.} an anomaly, is also a potential obstacle.
From here on, we will use the term anomaly detection to refer to generic obstacle detection, where the object category is not known at training time.
Beyond the ability to detect generic obstacles, another advantage of our approach is that the training does not require real or artificially created examples of obstacles on railways.
While visual anomaly detection has been extensively studied in the context of industrial inspection~\cite{Bergmann, Bergmann2019, Bergmann2020} and autonomous driving~\cite{Blum2021a, Yao2019}, Boussik \etal~\cite{Boussik2021} are the only ones addressing the problem of anomaly detection for railway environments.
They proposed training a series of \acp{AE} and using their reconstruction errors as a metric for anomalies.
In our experiments, we find that using reconstruction errors for obstacle detection fails, for example, for small obstacles or obstacles with colors that match the background.

We propose a novel data-driven approach to anomaly detection in railway environments by reformulating the problem as a local segmentation task, where the global information available to the network is restricted.
This bottlenecking limits over-confident predictions in the very well-defined and structured railway environment.
We train a network with a limited \textit{receptive field} (the patch size around a pixel the network sees) to segment railways from the background.
We also include random (non-railway) images as negative examples in the training procedure.
Inspired by Boussik \etal~\cite{Boussik2021}, we also study the incorporation of limited global information through obstacle-free images hallucinated by a neural network.
Our models are trained on an obstacle-free training set consisting of a subset of \textit{RailSem19}~\cite{Zendel2019} and evaluated on an obstacle-enhanced version of it (see Figure \ref{fig:teaser}), following a procedure proposed by Blum \etal~\cite{Blum2021a}.
%
%
We summarize our contributions as follows:
\begin{itemize}
    \item A novel approach using shallow networks to perform visual anomaly detection that is ideal for the highly structured environment of railways. Additionally, our method does not require hard-to-obtain images of real anomalies on railways.
    \item We explore the inclusion of global information through the use of hallucinated obstacle-free reconstructions and reformulate the anomaly detection problem as a semantic difference detection task.
    \item An extensive evaluation on an object-enhances version of the \textit{RailSem19} dataset~\cite{Zendel2019} comparing our solution to multiple baselines and evaluating the advantages and disadvantages of each method.
\end{itemize}

\section{RELATED WORK}
\label{sec:related_work}

\subsection{Visual Anomaly Detection}

In our case, visual anomaly detection is the task of detecting abnormalities or unknown entities in images based on the expected situation of the environment for normal operations.
According to Yang \etal~\cite{Yang2021}, works on visual anomaly detection can be classified into five categories.
\textbf{Probabilistic methods} such as Gaussian (mixture) models, kernel density estimation~\cite{Bharadwaj2021}, or variational \acfp{AE}~\cite{Kingma2014} aim at estimating the probability distribution of normal images and detecting the ones that fall out of the distribution.
\textbf{One-class classification methods} follow a similar approach by constructing a decision boundary either through support vector machines~\cite{Scholkopf2001}, support vector data descriptors~\cite{Tax2004} or deep learning~\cite{Oza2019, Ruff2018}. 
\textbf{Reconstruction-based methods} commonly use \acp{AE} to learn a low-dimensional representation from which the original input can be reconstructed~\cite{Bergmann2019, Akcay2019, Schlegl2017, Zenati2018}.
Reconstruction errors at test time are then used as a metric for anomalous image regions.
\textbf{Self-supervised anomaly detection methods} aim to learn significant and high-level features of normal samples, for example, through an auxiliary learning task~\cite{Golan2018, Lis2019}.
This avoids the need to precisely define what an anomaly is and does not require example data of anomalies.
\textbf{Feature modeling methods} do not detect anomalies in the image space, but in a hand-crafted or learned feature space, such as features of pre-trained \acp{CNN} \cite{Shi2021, Bergmann2020, Cohen2020}.
A good example is the student teacher distillation framework proposed by Bergmann \etal~\cite{Bergmann2020}, which achieved very good results on industrial images from the \textit{MVTec AD} anomaly detection dataset~\cite{Bergmann}.
The utilization of both deep and shallow \acp{CNN} has been explored for various of these methods~\cite{ruffUnifyingReviewDeep2021}, and it has been shown that shallow methods can outperform their deep counterparts in certain scenarios~\cite{kwonEmpiricalStudyNetwork2018}. 

\begin{figure*}[!t]
\centering
\includegraphics[width=.9\textwidth]{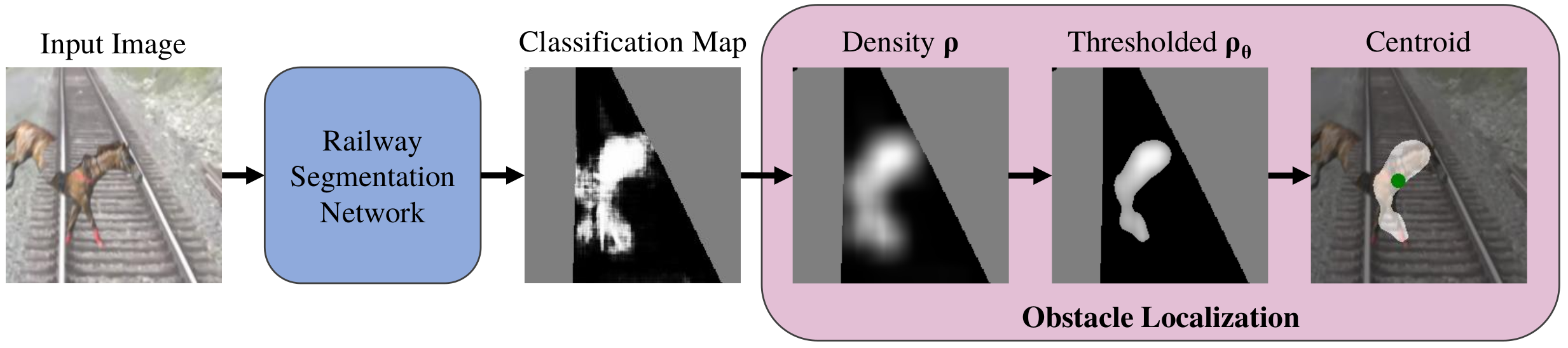}
\caption{Visual anomaly detection and localization using local segmentation. As input, we have a railway image with a synthetically added object. In the resulting classification map, the intensity of white represents the confidence that a pixel is anomalous, black denotes the railway, and the gray borders represent the background which is ignored. The final output is a detected object mask and its centroid.}
\label{fig:system-direct}
\end{figure*}

\subsection{Visual Obstacle Detection on Railways}

When developing new railway obstacle detection systems, it is important to consider the highly structured nature of the railway environment.
The unique shape and color of rails and rail ties, the well-defined trajectories, and the limited number of object categories on the train tracks facilitate the use of classical image processing methods.
This contrasts the more complex case of anomaly detection on roads, where learning-based methods are more essential~\cite{Yao2019}.
Ristić-Durrant \etal~\cite{ristic-durrantReviewVisionBasedOnBoard2021} have performed an extensive review of existing obstacle detection systems with a wide variety of sensors. 
We focus on camera-based systems, which are directly comparable to our proposed approach.
Rüder \etal~\cite{Ruder2003} introduce a system for track and obstacle detection using edge detection, optical flow, and statistics of texture in an approach that is heavily tailored towards domain-specific grayscale images.
However, evaluations regarding different object categories remain limited. 
Mukojima \etal~\cite{mukojimaMovingCameraBackgroundsubtraction2016, nakasoneFrontalObstacleDetection2017} propose a background subtraction-based method, which is, however susceptible to changes in lighting or environment. 
Rodriguez \etal~\cite{Rodriguez2012} address this problem on the railway tracks themselves, which are observed using a Hough transform and a Canny edge detector.
Discontinuities in this detection process signify obstacles on the track, though with limited robustness, as assumptions about the track geometry cause this approach to fail in scenarios with curved tracks.
Uribe \etal~\cite{uribeVideoBasedSystem2012} follow a similar approach with the same shortcomings.
Learning-based object detection methods have also shown some success in detecting humans, trains, or luggage~\cite{Yu2018, liRealWorldRailwayTraffic2018, Xu2019, Chernov2020, Ristic-Durrant2021a,durrant2022, mahtaniPedestrianDetectionClassification2020, guptaDeepVisionbasedSurveillance2022, yeRailwayTrafficObject2021}, but they rely on custom datasets and are limited to a fixed set of object categories.

To this point, there is only little research applying visual anomaly detection methods to railway obstacle detection. 
Gasparini \etal~\cite{Gasparini2020} combine unsupervised image reconstruction with supervised detection of anomalies for nighttime railway inspection but are thus limited to thermal cameras.
Boussik \etal~\cite{Boussik2021} perform a grid search over \ac{AE} structures with different optimizers, activations, and loss functions and evaluate them on a custom test dataset with artificially inserted obstacles and one real-world scenario.
In our experiments, these \ac{AE}-based methods fail to detect small obstacles or those with colors common in the railway environment.
Wang \etal~\cite{Wang2022} similarly train an \ac{AE}, but detect anomalies by directly analyzing the distribution in the latent space and solely utilize the reconstruction for localizing detected anomalies.

\section{Anomaly Detection Through Local Segmentation}
\label{sec:method_segmentation}

A challenge for data-driven methods is the lack of existing training data featuring obstacles on railways and the difficulty of obtaining these images in such a safety-critical environment.
This limits us to obstacle-free railway images and to other datasets featuring possible obstacles in non-railway scenarios.
Additionally, we cannot limit detection to a fixed set of object categories, as we cannot predict in advance all possible obstacles we might encounter.
Our approach to visual anomaly detection allows us to detect obstacles implicitly by exclusion.
If it is on the railway but does not conform to known railway patterns, \textit{i.e.} an anomaly, it is a potential obstacle.

We train an anomaly detection network by segmenting railways from the background as an auxiliary task.
What sets our method apart is how we exploit the very structured and repetitive nature of our environment.
Railway tracks have a very distinctive structure, which is easily identifiable even when looking at small patches taken from a larger image.
We use a network with a small \textit{receptive field} (the size of the patch around a pixel the network sees) as a means to restrict the global (\textit{i.e.} contextual) information we provide to the anomaly detection network.
This information bottleneck prevents issues of overconfidence in segmentation and classification tasks~\cite{gastLightweightProbabilisticDeep2018a}, examples of which we also show later in Section~\ref{sec:results_fishyrails}.

To provide negative examples, we include random (non-railway) images featuring a large variety of objects and scenes.
Note that, in contrast to supervised object detection, we do not take an opinion on the semantic classes of obstacles but we label the entire image as background.
The network is trained by minimizing the \ac{BCE} loss $\mathcal{L}_{BCE}$ for every pixel.
We exclude the loss outside the track masks for railway images to avoid boundary issues and mislabeled pixels.
Thus, we let all the negative examples come from non-railway images.

\subsection{Obstacle Localization With Classification Maps}
\label{sec:method_localization}

\begin{figure}[t]
\centering
\includegraphics[width=0.8\linewidth]{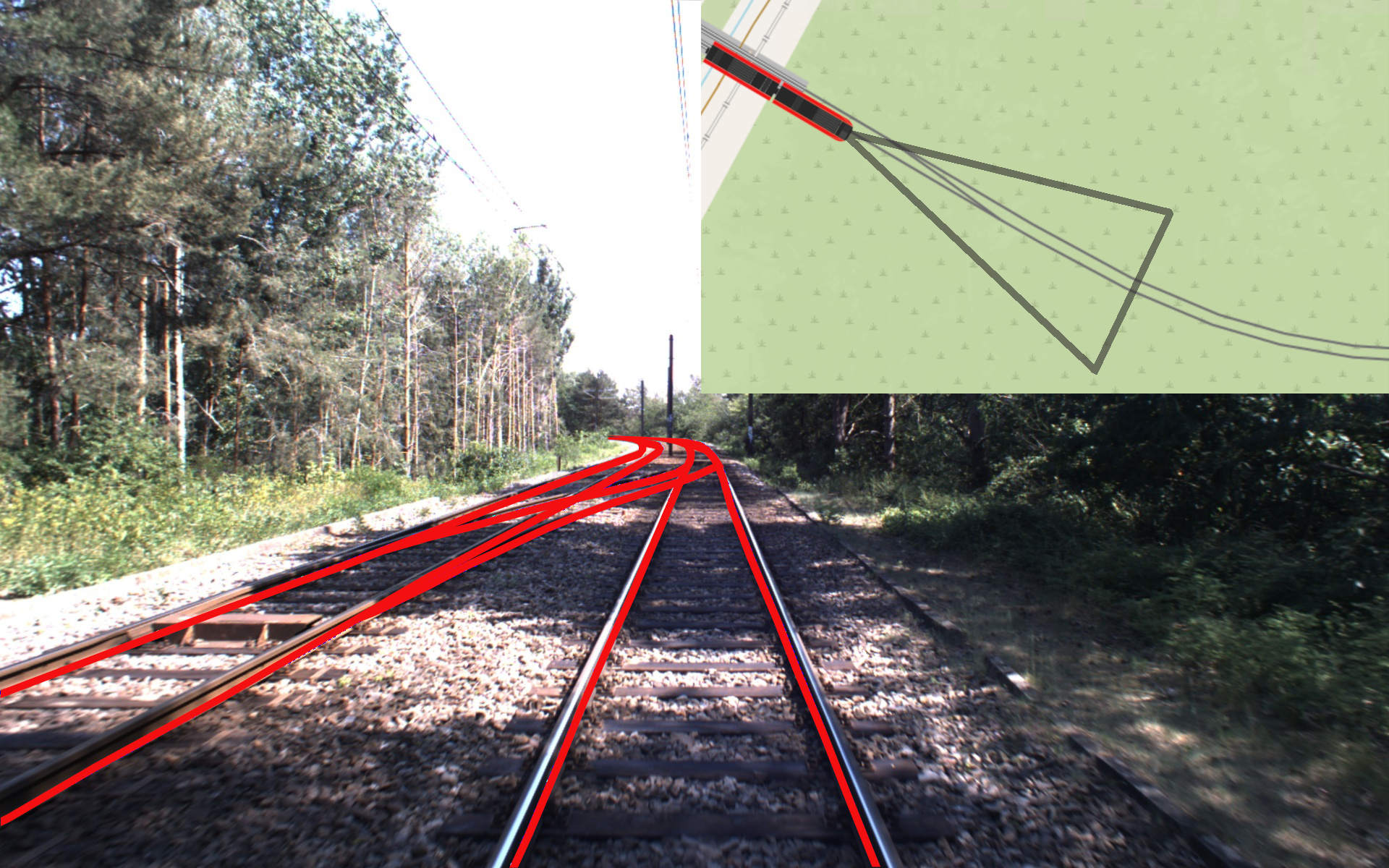}
\caption{Railway tracks being re-projected into the camera frame based on the known location and orientation of the train from an RTK-GPS.}
\label{fig:map-reprojection}
\end{figure}

An overview of the system is shown in Figure~\ref{fig:system-direct}.
During deployment, the previously trained network provides a classification map $\{$\texttt{railway}, \texttt{background}$\}$ covering each pixel in the image.
Anomalies are then pixels that we know are railway but are classified as background.
However, this requires a ground truth rail track segmentation mask to compare against.
We assume prior knowledge about the location of the railway tracks in the image, which in practice can be obtained by having a prior map of the railway network and knowing the current pose of the train~\cite{huFusionVisionGPS2006, tschoppExperimentalComparisonVisualAided2019, oteguiSurveyTrainPositioning2017}.
Using the map, the position of the train, and the camera intrinsics, the rail map can be projected into the image frame to obtain a ground truth mask as shown in Figure~\ref{fig:map-reprojection}.

From a given classification map, obstacles are localized by computing a density map $\bm{\rho}$ via uniform filtering with a filter size of $K$.
Subsequently, a threshold $\theta$ is applied, where all pixel values smaller than $\theta$ are zeroed to obtain the final obstacle mask $\bm{\rho_\theta}$.
Decisions on the existence of the object can then be made based on the number of anomalous pixels or the size of the connected anomalous regions, as obtained, for example from a clustering algorithm.

\section{Incorporating Global Information}
\label{sec:method_global}

\begin{figure}[!t]
\centering
\includegraphics[width=.9\linewidth]{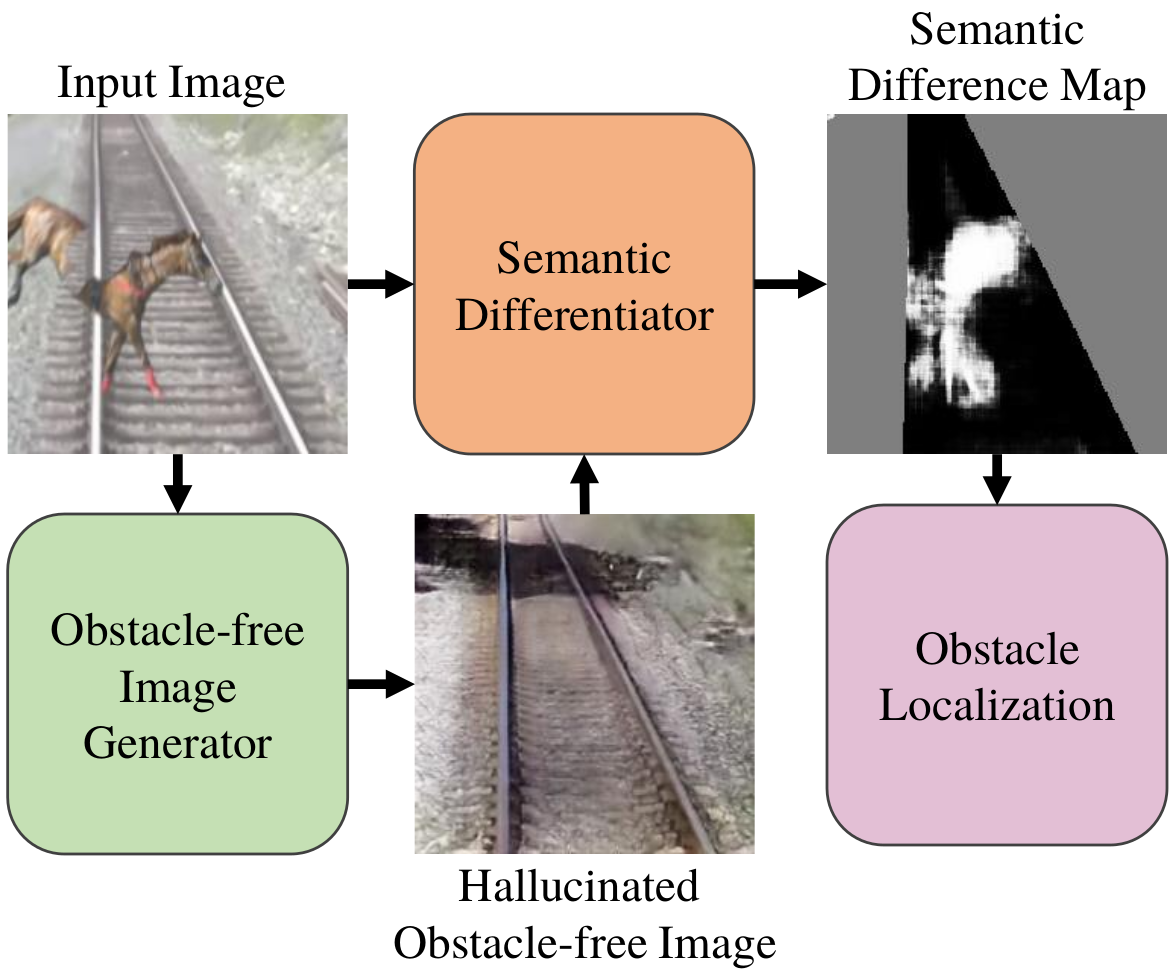}
\caption{Overview of the obstacle detection using hallucinated obstacle-free images. A semantic differentiator network does a pixel-wise prediction of where the original and obstacle-free images have different semantic meanings. This difference map can then be used to localize anomalies, as shown in Figure~\ref{fig:system-direct}.}
\label{fig:system-gan}
\end{figure}

We hypothesize that incorporating global information into our, so far, purely local approach would lead to better performance.
Inspired by the reconstruction-based approaches by Boussik \etal~\cite{Boussik2021}, we use a neural network to hallucinate obstacle-free images.
However, instead of looking at the reconstruction error, we compute the semantic class difference between regions in the original image and the hallucinated obstacle-free reconstruction (see Figure~\ref{fig:system-gan}).
The semantic differentiation network predicts in which pixels in the two images the semantic class is different without ever having to explicitly predict the class.
For the same reasons as stated in Section~\ref{sec:method_segmentation}, we design the semantic differentiator network with a limited receptive field.
This allows us to maintain a local informational bottleneck, which we have seen to be beneficial, while also allowing an earlier network to incorporate global information.

\subsection{Obstacle-Free Railway Image Generation}
For generating synthetic, obstacle-free railway images, we train the network with an \acf{AE} structure to reconstruct the input image through a low-dimensional bottleneck.
Choosing a small bottleneck prevents the network from simply replicating its input and forces the decoder to encode in its weights repeating patterns in the training data.
As the training data does not include obstacles, the decoder should never learn how to reconstruct them, and the resulting images should be obstacle-free versions of the original input~\cite{Boussik2021}.
We test in total four different combinations of reconstruction losses that enforce different properties.

\textit{1)} The \textbf{\ac{MSE}} loss $\mathcal{L}_{MSE}$ computes the mean per-pixel $L_2$ distance between the original and reconstructed image.
\ac{MSE} pushes the network's output toward the dataset average, which causes the network to only preserve low-frequency components in the image at the cost of finer structural details.
%

\textit{2)} The \textbf{\ac{SSIM}} loss $\mathcal{L}_{SSIM}$ as proposed by Wang \etal~\cite{Wang2004} forces patches in the original and reconstructed image to have similar luminance, contrast, and structure.
This results in sharper images, but still, the network can only preserve a limited amount of realism in the reconstructed image.
%

\textit{3)} Images generated from the previous methods can easily be identified as synthetic, which leads us to \textbf{\acp{GAN}} for more realistic image generation~\cite{Goodfellow2014, Luo2019}.
Inspired by Isola \etal~\cite{Isola2017}, we use a conditional \ac{GAN} architecture~\cite{Mirza2014} with a loss $\mathcal{L}_{GAN}$ to generate obstacle-free railway images.
In an adversarial setting, the auto-encoding generator's objective is to reconstruct realistic (obstacle-free) railway images, while a discriminator aims at distinguishing real images from fake generated ones.
Both are conditioned using our original segmentation mask, to promote a semantic similarity and to conserve the rail trajectories.
Important to note is that apart from a semantic similarity constraint, the adversarial loss $\mathcal{L}_{GAN}$ does not promote visual similarity, such as color or structure, between the two images.

\textit{4)} To preserve visual similarity, we expand the adversarial loss by applying the two \textbf{Histogram Losses} proposed by Avi-Aharon \etal~\cite{Avi-Aharon2020}.
As suggested by the authors, we combine both their proposed mutual information and Earth Mover's Distance losses as $\mathcal{L}_{HIST}$ into the \ac{GAN} training setup.
These two loss functions aid the \ac{GAN} in retaining structural and color information, respectively, instead of pure semantic similarity.

\subsection{Object Localization with Semantic Difference Maps}
\label{sec:method_diff}

The image reconstructed by our network provides global information about how each pixel would look like in an obstacle-free image.
We leverage this information by training a network to compute the pixel-wise semantic difference map between the original and reconstructed obstacle-free image, as shown in Figure~\ref{fig:system-gan}.
%
%
We train the network with semantically similar and different image pairs. 
In the semantically similar case, we choose original and hallucinated versions of the same obstacle-free railway image and label every pixel as similar.
In the semantically different case, we simulate an obstacle by replacing the original image with a random image from a dataset of non-railway images as in Section~\ref{sec:method_segmentation} and label every pixel as different, thus not focusing on specific classes.
As before, we train the network by minimizing the \ac{BCE} loss for every pixel and ignoring pixels outside the ground truth railway. 
Obstacles can then be localized using the same procedure as in Section~\ref{sec:method_localization} and Figure~\ref{fig:system-direct} on the difference map instead.

\begin{figure*}[t]
    \centering
    \includegraphics[width=0.9\linewidth]{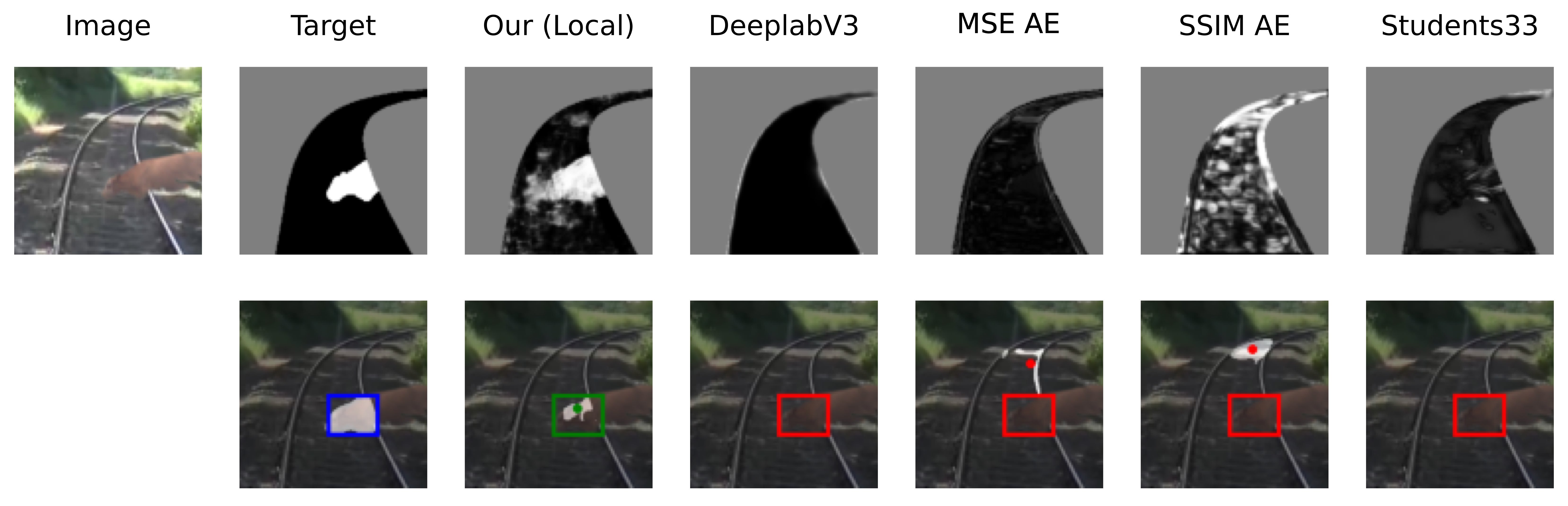}
    \caption{Example image where only our local method localizes the obstacle correctly, while all baseline methods fail. 
    The first row shows the predicted obstacle segmentation maps.
    Gray represents the areas outside the railway tracks where no detections are done, in black is the railway and white represents the anomaly detection mask.
    The second row displays localization results obtained via centroid computation.}
    \label{fig:results_patchclass}
\end{figure*}

\section{EXPERIMENTAL SETUP}
\subsection{Datasets}
\label{sec:experiments_datasets}

As a training dataset, we use the annotated subset of 7500 training images from the public \textit{RailSem19} dataset~\cite{Zendel2019}, excluding tramway images.
From these images, various regions of interest crops are obtained at a resolution of $224 \times 224$, resulting in a total of $26,810$ images.
As additional non-railway images $\mathcal{X}_I$, we take $1,281,167$ crops from the public~\textit{ImageNet} dataset~\cite{Deng2009}.

For our evaluation dataset \textit{FishyRails}, we take the official annotated test set of \textit{RailSem19}.
We populate the images with objects from the public \textit{PascalVOC} object segmentation dataset~\cite{Everingham2015}, resulting in $7142$ images and segmentation masks.
The pasting process is inspired by \textit{Fishyscapes}~\cite{Blum2021a}, an object-enhanced dataset for measuring segmentation blind spots in traffic images, and involves image border smoothing, brightness correction, motion blur, depth blur, and Gaussian noise. 
Similar augmentations have also been proposed by~\cite{s19143075, mukojimaMovingCameraBackgroundsubtraction2016, uribeVideoBasedSystem2012, durrant2022} for use in railway anomaly detection, as there are no public real-world railway obstacle detection datasets.
Our synthetic obstacle testing method is also in line with the one proposed by Boussik \etal~\cite{Boussik2021}, who instead use a \ac{GAN} to blend images of railways and obstacles on the \textit{RailSem19} dataset.

\subsection{Baselines}
\label{sec:experiments_baselines}

We compare against a set of state-of-the-art baselines selected from the related works. 
A standard semantic segmentation network, \textit{DeeplabV3}~\cite{Chen2017}, trained on obstacle-free railway images, to simply differentiate railway from background, serves as a first baseline.
The second and third baselines are based on the work of Boussik \etal~\cite{Boussik2021}, consisting of two \acfp{AE} for railway images, trained with either a \acf{MSE} or \acf{SSIM} loss.
The reconstruction error between the output and the original image is then used as a measure of anomalousness.
We call these two baselines \textit{MSE AE} and \textit{SSIM AE} respectively.
As our last baseline, we consider a patch-wise student teacher method\footnote{Our re-implementation, as the original source code is not available.} proposed by Bergmann \etal~\cite{Bergmann2020}.
The authors report impressive results for visual anomaly detection on the \textit{MVTec AD} industrial image dataset~\cite{Bergmann}, and we expect comparable results in our uniform and equally well-structured railway environment.

\subsection{Metrics}

In order to assess the quality of the classification or difference map, we use the \acf{AUROC}. 
The \ac{AUROC} is computed over all pixels of interest in the dataset given a ground-truth segmentation mask.
Even though this is a good threshold-independent metric, it does not reflect well the applicability for our target use case of obstacle localization.

We say an obstacle is correctly localized if and only if the predicted centroid lies within the bounding box of the ground truth obstacle.
Based on this localization, the F1 score can be computed as a second evaluation metric.
To compute the centroid, we calculate the mean of the coordinates of all non-zero pixels (as seen in Figure~\ref{fig:system-direct}).
For all methods and baselines, we individually grid search for the optimal $K$ and $\theta$ (see Section~\ref{sec:method_localization}) that maximize the F1 score.
To note is that we implicitly assume there is at most one object in the image.
This assumption holds in our experiments and provides a useful comparison metric.
In practice, clustering could be used to distinguish multiple objects or an alert could be triggered based on the amount of anomalous pixels.

\section{Results}
\label{sec:results}

\subsection{Obstacle detection on \textit{FishyRails}}
\label{sec:results_fishyrails}

\begin{table}[!t]
\begin{center}
 \caption{
 The \ac{AUROC} and localization F1 score on \textit{FishyRails}.
 }
 \vspace{1ex}
 \label{tab:results}
 \begin{tabular}{ll|c|c}
 \toprule
 Method & \ac{AE} loss & \ac{AUROC} & F1 \\ \midrule
 DeeplabV3~\cite{Chen2017} & - & 0.817 & 0.535 \\
 MSE AE~\cite{Boussik2021} & $\mathcal{L}_{MSE}$ & 0.686 & 0.498 \\
 SSIM AE~\cite{Boussik2021} & $\mathcal{L}_{SSIM}$ & 0.737 & 0.429 \\
 Students33~\cite{Bergmann2020} & - & 0.594 & 0.458 \\ \midrule
 \textbf{Our} (Local) & - & 0.926 & \textbf{0.863} \\
 \textbf{Our} (Global) & $\mathcal{L}_{MSE}$ & 0.917 & 0.825 \\
 \textbf{Our} (Global) & $\mathcal{L}_{SSIM}$ & 0.915 & 0.812 \\
 \textbf{Our} (Global) & $\mathcal{L}_{GAN}$ & 0.935 & 0.857 \\
 \textbf{Our} (Global) & $\mathcal{L}_{GAN} + \mathcal{L}_{HIST}$ & \textbf{0.936} & 0.838 \\
 \bottomrule

 \end{tabular}
\end{center}
\end{table}

\begin{figure*}[t]
    \centering
    \includegraphics[width=0.8\linewidth]{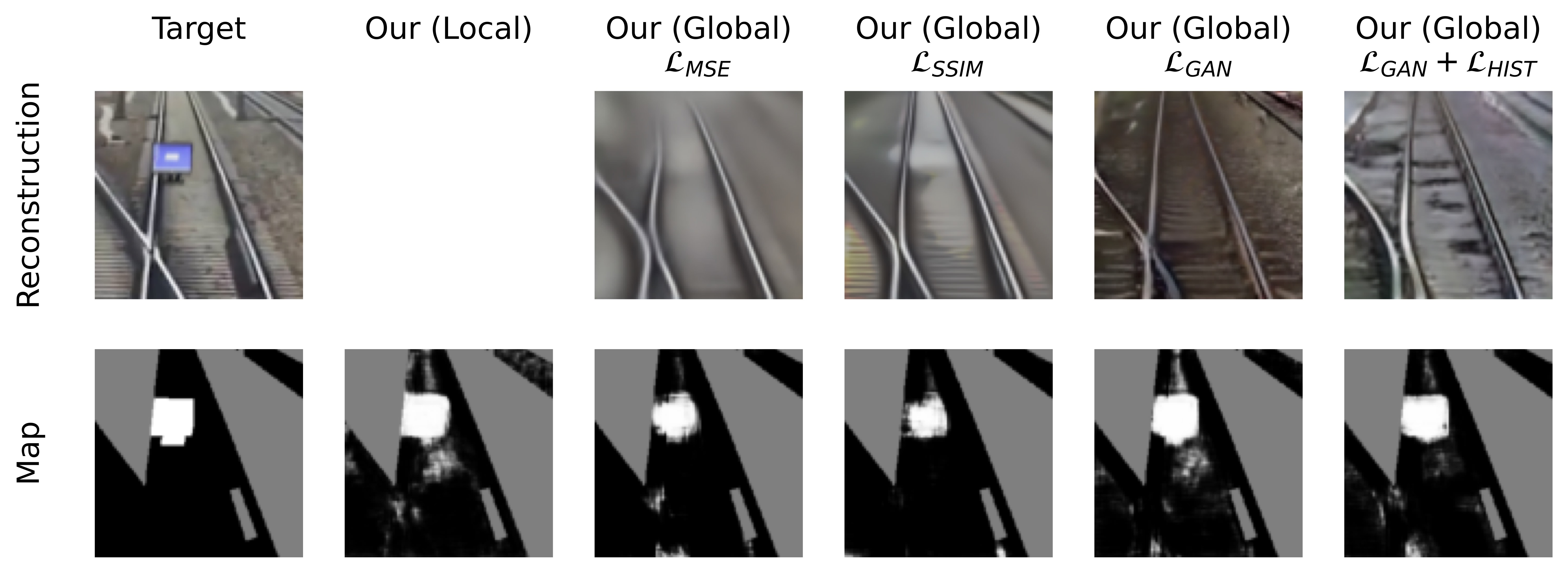}
    \caption{Example image where both our local and global methods correctly segment the obstacle. 
    The obstacle-free reconstructions are compared semantically with the original image on the left to find differences that could be anomalies.
    Note that the reconstructed images look different depending on the loss used to train the obstacle-free image generator.
    Nevertheless, in this example, all reconstructions successfully ignore the obstacle. }
    \label{fig:results_patchdiff1}
\end{figure*}

For each method separately we use a fixed $\rho$ and $\theta$ (see Section~\ref{sec:method_localization}) across all experiments.
We determine the two thresholds separately for each method by picking the best-performing value across the training set.

After evaluation on our \textit{FishyRails} dataset, the \ac{AUROC} and F1 scores, as reported in Table \ref{tab:results}, paint a clear picture: our methods outperform the baselines by a large margin. 
Both the local and best global version ($\mathcal{L}_{GAN} + \mathcal{L}_{HIST}$) of our proposed method achieve an \ac{AUROC} of 0.926 or better.
This is significantly better than the \ac{AUROC} of the best baseline method (\textit{DeeplabV3}) at 0.817. 
The difference in F1 score is even larger, as our purely local method achieves a score of 0.863, while \textit{DeeplabV3} only reaches a value of 0.535. 
We highlight this also visually in Figure~\ref{fig:results_patchclass} which shows an example where our local method succeeds at detecting the obstacle, whereas baseline methods fail.
The comparison to \textit{DeeplabV3} is important as it mirrors our local approach, except for a deeper network and a much higher receptive field.
This highlights the importance of reducing the receptive field of the network to restrict global information and experimentally validates our hypothesis from Section~\ref{sec:method_segmentation}.

\begin{figure}[t]
    \centering
    \includegraphics[width=\linewidth]{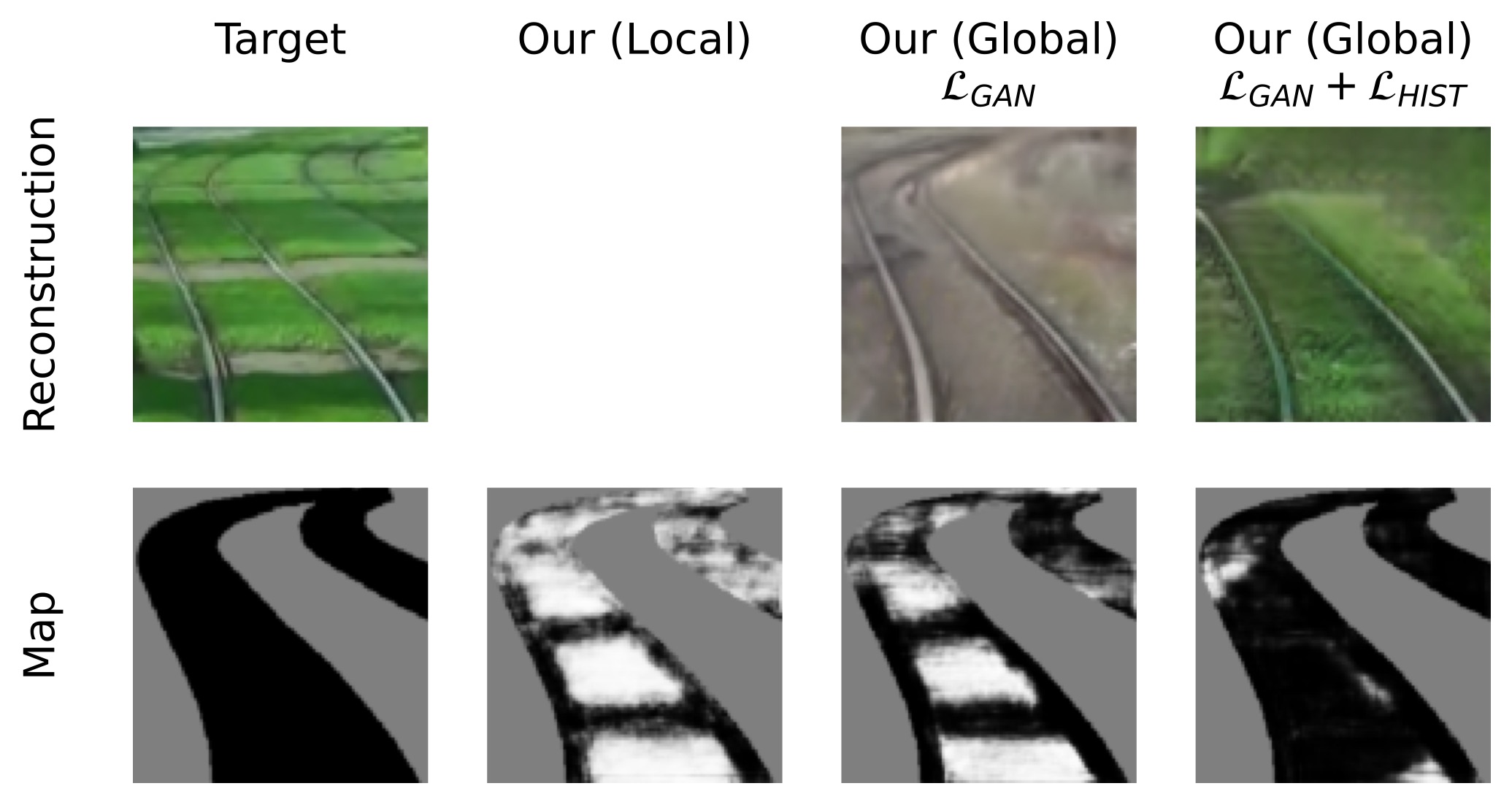}
    \caption{Our global method with $\mathcal{L}_{GAN} + \mathcal{L}_{HIST}$ successfully reconstructs grass as non-anomalous. This allows the semantic difference network to correctly predict the images are similar and that there is no obstacle. The simpler \ac{GAN} succeeds in reconstructing the railway but fails to match colors as it has not seen enough examples of grassy railways during training.}
    \label{fig:results_patchdiff2}
\end{figure}

Interestingly, we find that \textit{DeeplabV3} outperforms both the student teacher method \textit{Students33}~\cite{Bergmann2020} and the reconstruction-based methods\textit{MSE AE}~\cite{Boussik2021} and \textit{SSIM AE}~\cite{Boussik2021}.
\textit{DeeplabV3} tends to over-confidently classify obstacle pixels as railway.
During training, the model seems to learn that areas in between train tracks tend to correspond to the railway class, as it has never seen obstacles during training. 
This leads to small objects or ones with similar color as the background being misclassified.

From our investigations, we find that both \textit{MSE AE} and \textit{SSIM AE} fail when obstacles are small or have similar color as the background, with \textit{SSIM AE} performing slightly better on obstacles with very prominent structure and contrast.
Among all methods, \textit{Students33} produces the lowest \ac{AUROC} and F1 scores, probably because it was designed and optimized for industrial images with even less variance in structure and color than our dataset.
The student networks are capable of detecting obstacles with salient colors but are unable to detect obstacles of colors that were frequently observed during student training, such as brown or gray.

\begin{figure}[t]
    \centering
    \includegraphics[width=\linewidth]{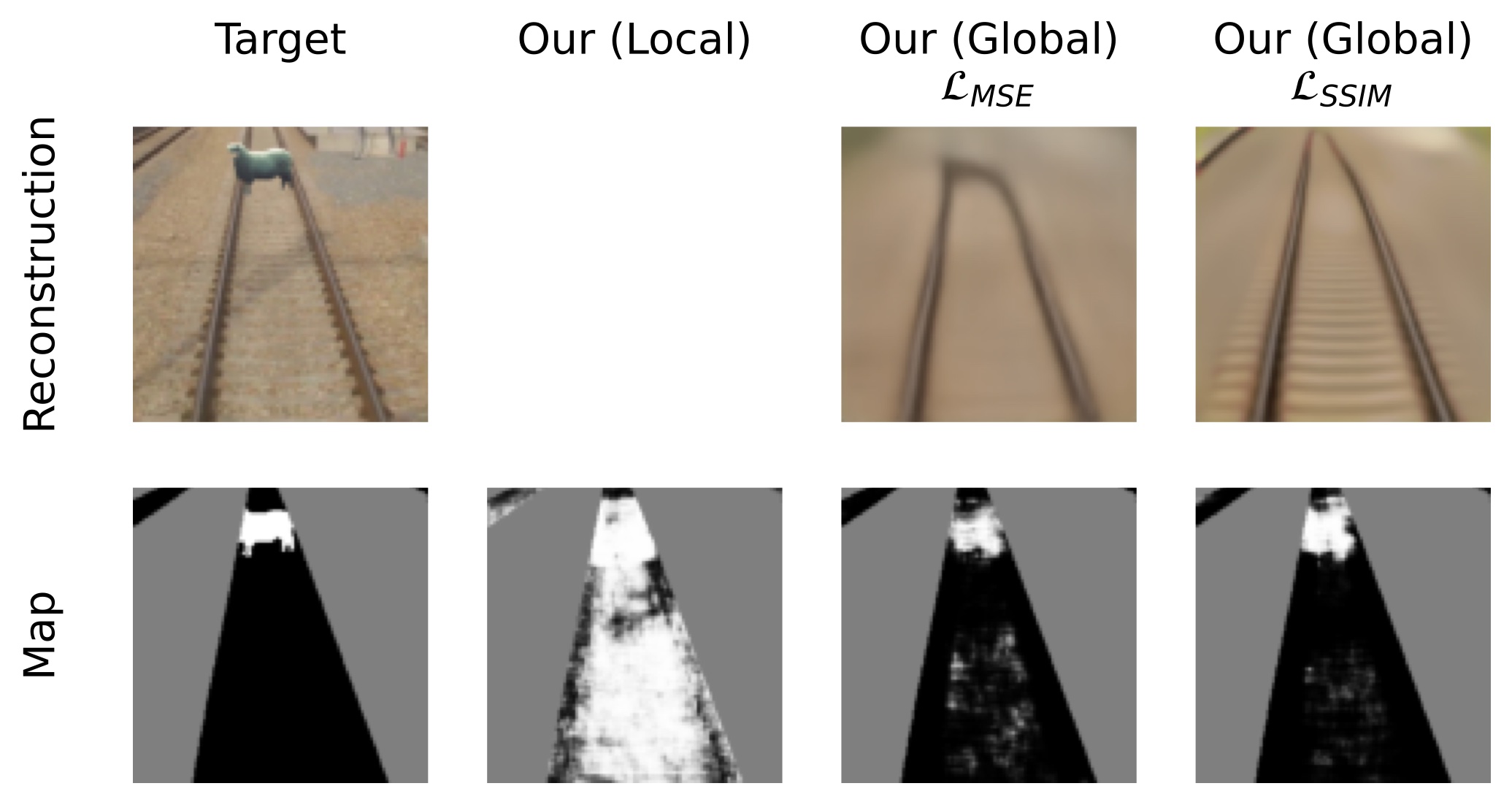}
    \caption{Our local method misclassifies the sandy background as anomalous, because of lack of sand between the tracks during training. However, the $\mathcal{L}_{MSE}$ and $\mathcal{L}_{SSIM}$ networks succeed in removing the obstacle because of the distinct color difference to the railway.}
    \label{fig:results_patchdiff4}
\end{figure}

\subsection{Should We Include Global Information?}
\label{sec:results_global}

While the results reported in Table~\ref{tab:results} show that our methods outperform the baselines by a large margin, the difference between our local (Section~\ref{sec:method_segmentation}) and global (Section~\ref{sec:method_global}) methods, is much smaller.
The best \ac{AUROC} score of 0.936 is achieved by our global method with $\mathcal{L}_{GAN} + \mathcal{L}_{HIST}$, but our local method yields the best F1 score.
Therefore, a more detailed comparison of the observed strengths and failure cases in our experiments is needed. 
All global methods manage to remove most of the obstacles in their reconstruction stage for the four different loss combinations we tested (see Figure~\ref{fig:results_patchdiff1} for examples).
When trained on anomaly-free railway and random non-railway patches, our generative models trained with $\mathcal{L}_{MSE}$ and $\mathcal{L}_{SSIM}$ learn to focus on pixel-wise color differences instead of semantic differences between the original and generated image.
This leads to problems where small obstacles or ones with similar background colors are reconstructed too well, instead of being removed.

\begin{figure}[t]
    \centering
    \includegraphics[width=\linewidth]{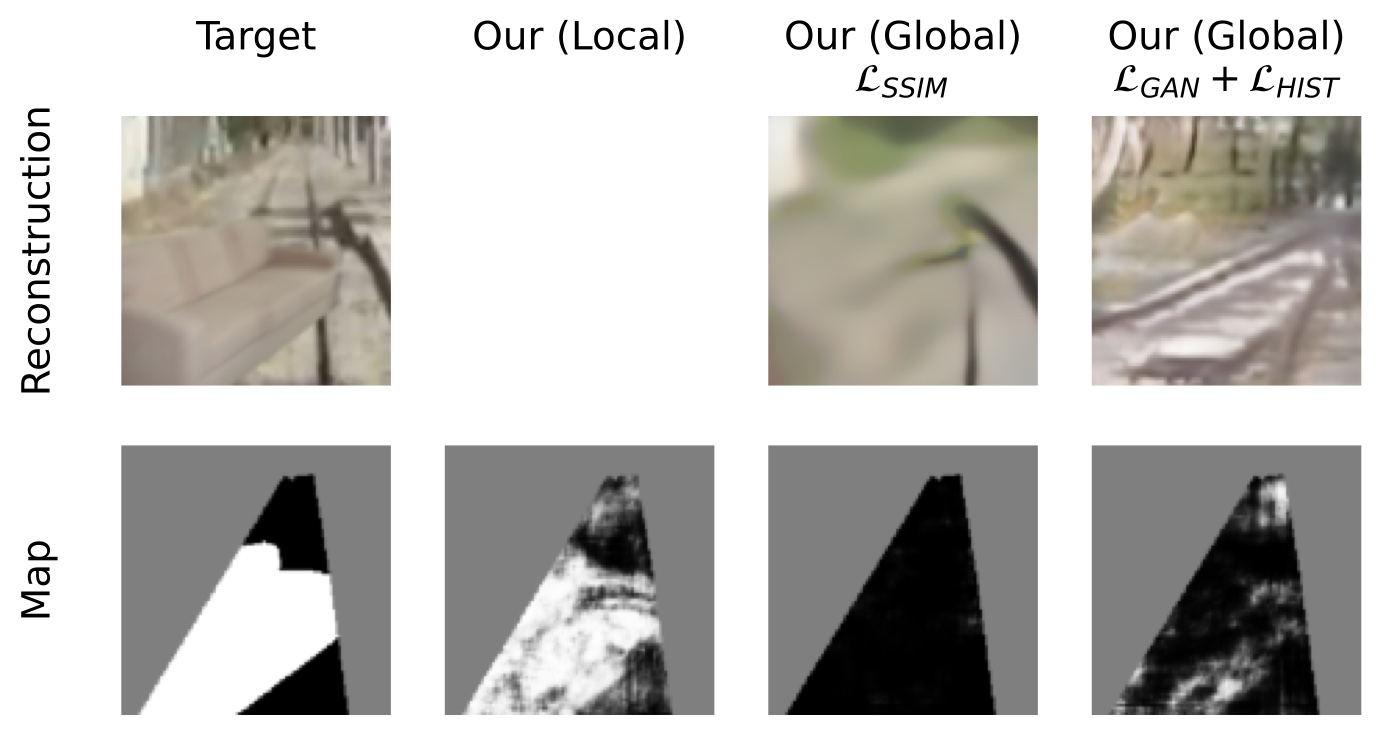}
    \caption{Our global methods fail to reconstruct obstacle-free images with large occlusions, as they provide no prior on the potential tracks. The reconstruction failure also leads to the semantic difference map being meaningless. Our local method succeeds as it does not depend on the global context.}
    \label{fig:results_patchdiff3}
\end{figure}

The \ac{GAN} based methods with losses $\mathcal{L}_{GAN}$ and $\mathcal{L}_{GAN} + \mathcal{L}_{HIST}$ succeed at rarely observed scenes by focusing on both visual and semantic reconstruction (see Figures \ref{fig:results_patchdiff2} and~\ref{fig:results_patchdiff4} for examples).
This leads to fewer false positives and is an instance where the inclusion of some global information into the process outperforms our purely local anomaly detection network.
In cases where obstacles are too large, \textit{i.e.} they cover too much of the image frame, the obstacle-free image generation process fails entirely, as seen in Figure~\ref{fig:results_patchdiff3}. 
This is an instance where our purely local method outperforms global methods, as it does not rely on a reconstruction stage.
Overall, the local and global versions of our methods have different weaknesses and strengths, while having similar overall performance according to our metrics.
Ideally, a combination of multiple classifiers could be used.
However, in practice, the local method is simpler and thus has fewer failure modes and might be preferable in a safety-critical application.

\subsection{Ablation Study}
\label{sec:abl_study}
We perform an ablation study on the receptive fields of both our local and global approaches using different loss functions. 
The results for the local method and the global ($\mathcal{L}_{GAN} + \mathcal{L}_{HIST}$) method are shown in Table~\ref{tab:results_patch} and highlight that according to the F1 score, a receptive field of $\unit[21]{px}$ is the optimum for us and has therefore been used in the evaluations of our method.
For our global method, a similar pattern can be observed with a maximum at a receptive field of $\unit[29]{px}$ and utilizing the $\mathcal{L}_{GAN} + \mathcal{L}_{HIST}$ loss.

\begin{table}[t!]
\begin{center}
\caption{Ablation study on the size of the receptive field $K_p$ [pixels] of our local and global method utilizing the $\mathcal{L}_{GAN} + \mathcal{L}_{HIST}$ loss.}\vspace{1ex}
\label{tab:results_patch}
\begin{tabular}{ll|c|c}
\toprule
Method & $K_p$ & ROC AUC & F1\\ \midrule
\textbf{Our} (Local) & 13 & 0.921 & 0.836 \\ 
\textbf{Our} (Local) & 21 & 0.926 & \textbf{0.863} \\
\textbf{Our} (Local) & 29 & \textbf{0.928} & 0.861 \\
\textbf{Our} (Local) & 35 & 0.925 & 0.839 \\
\textbf{Our} (Local) & 51 & 0.927 & 0.832 \\
\midrule
\textbf{Our} (Global) & 13 & 0.931 & 0.839 \\
\textbf{Our} (Global) & 21 & \textbf{0.936} & 0.838 \\
\textbf{Our} (Global) & 29 & \textbf{0.936} & \textbf{0.846} \\
\textbf{Our} (Global) & 35 & 0.930 & 0.826 \\
\textbf{Our} (Global) & 51 & 0.914 & 0.782 \\
\bottomrule
\end{tabular}
\end{center}
\end{table}

\section{CONCLUSION}
In this paper, we have presented a novel approach to data-driven obstacle detection on railways. 
By training a network with a restricted receptive field on an auxiliary segmentation task, we are able to discern the well-structured railway background from any anomalies (obstacles).
We are able to train the network without the need for hard-to-obtain data of obstacles on railways, and without having to restrict ourselves to a limited set of obstacle classes.
Our method succeeds in cases where the benchmarks fail, \textit{e.g.} small obstacles or obstacles with colors that blend into the background, but struggles with rarely seen railway environment types.
In an extension, global information can be incorporated through hallucinated obstacle-free reconstructions. 
Successful reconstructions help detect anomalies even in rarely seen environments, though if the reconstruction fails, also the detection itself fails.
Due to the limited availability of railway anomaly datasets, we evaluate our system on an obstacle-enhanced version of \textit{Railsem19}, showing a significant improvement over state-of-the-art baselines.
%


\begin{acronym}
    \acro{SSIM}{Structural Similarity}
    \acro{CNN}{Convolutional Neural Network}
    \acro{BCE}{Binary Cross-Entropy}
    \acro{MSE}{Mean Squared Error}
    \acro{GAN}{Generative Adversarial Network}
    \acro{MI}{Mutual Information}
    \acro{EMD}{Earth Mover's Distance}
    \acro{AUROC}{Area Under the Receiver Operator Characteristic Curve}
    \acro{RMSE}{Root Mean Square Error}
    \acro{ETCS}{European Train Control System}
    \acro{AE}{Auto-Encoder}
    \acro{Lidar}{Light detection and ranging}
\end{acronym}

\bibliographystyle{ieeetran}
\bibliography{root.bib}

\end{document}